# A Spatial-temporal 3D Human Pose Reconstruction Framework

Xuan Thanh Nguyen *, Thi Duyen Ngo ** and Thanh Ha Le **

**Abstract**
3D human pose reconstruction from single-view camera is a difficult and challenging topic. Many approaches have been proposed, but almost focusing on frame-by-frame independently while inter-frames are highly correlated in a pose sequence. In contrast, we introduce a novel spatial-temporal 3D reconstruction framework that leverages both intra and inter frame relationships in consecutive 2D pose sequences. Orthogonal Matching Pursuit (OMP) algorithm, pre-trained Pose-angle Limits and Temporal Models have been implemented. We quantitatively compare our framework versus recent works on CMU motion capture dataset and Vietnamese traditional dance sequences. Our method outperforms others with 10 percent lower of Euclidean reconstruction error and robustness against Gaussian noise. Additionally, it is also important to mention that our reconstructed 3D pose sequences are smoother and more natural than others.

## 1. Introduction

Reconstruction of 3D human pose from single-view camera plays an important role in many computer vision research fields such as animation, human-computer interaction and video surveillance applications [1]. A human action is naturally represented by a sequence of poses which is usually described by a hierarchy skeleton structure of joint positions. Finding 3D corresponding pose of 2D pose is obviously ambiguous since many plausible 3D poses satisfy a single 2D pose [2].

Previous approaches were introduced to reconstruct 3D human pose from monocular/multi-view image, video, depth channel. However, these approaches typically focus on single frame independently while all frames in a sequence are highly correlated [3] [4] [5] [6]. Others lay on fixed joint angle limit models which are not general enough for the diverse of human poses and sometimes leading to invalid reconstructed 3D poses [7] [8].

Our goal is taking advantages of frame correlation in a sequence of poses. In this paper, we focus on input data of 2D pose sequences, the key idea of our framework is combining intra and inter frames relationships in a consecutive pose sequence. Given a 2D human pose sequence, we firstly use Orthogonal Matching Pursuit algorithm and pre-trained Pose-angle Limits model to estimate prior 3D pose results. This pre-trained pose-dependent model was introduced by [3] that solved the fixed joint angle problems. Secondly, temporal models independently correct and smooth sequence of 3D poses to generate final 3D poses. Our method have been tested on CMU dataset and recorded Vietnamese Traditional Dance sequences to show state-of-the-art reconstruction results compare to other existing methods.

Our contribution are:
• Proposing a spatial-temporal 3D reconstruction framework that takes advantage of both intra and inter frame relationships in consecutive 2D pose sequences.
• Using temporal models as an post-processing step to smooth reconstructed motion sequences, this module can be re-used independently in other methods.

The remainder of this paper is organized as follows: Section 2 gives a short literature review of state-of-the-art 3D human pose estimation studies. Section 3 describes our proposed framework using Orthogonal Matching Pursuit (OMP) algorithm, pre-trained Pose-angle Limits and Temporal Models. Experimental results on CMU motion capture dataset and Vietnamese traditional dance sequences are given in Section 4. Finally, in Section 5, we briefly conclude our method.

## 2. Related Works

Fundamentally, 3D human pose reconstruction from single color camera is an under-constrained problem which is much more challenging than multi-view or depth channel reconstruction since lacking of three-dimensional information. Typically, 2D human poses are firstly estimated from image/video, then 2D-to-3D poses conversion is the second step. Within scope of this work, our paper specifically focuses on major contribution of 3D poses temporal model reconstruction from 2D poses. For the first step of singe-view image/video to 2D pose, this paper re-use Deepcut [9] which is a widely known deep learning-based framework for 2D pose detection.

Many approaches have been proposed to reconstruct 3D human pose from only monocular view data [3] [10] [11] or video sequences [4] [5]. Ramakrishna et al. [3] presented an activity-independent method to recover 3D configuration from a single image using a orthogonal matching pursuit (OMP) algorithm and 3D pose sparse representation on an over-completed dictionary of basic poses. They take advantage of a large motion capture corpus as a proxy for visual memory to draw a plausible 3D configuration. However, this model has problem with strong perspective effects image where weak perspective assumptions on the camera model are violated and mean pose is not a reasonable initialized. Tekin et al. [11] proposed a deep learning regression for structure prediction of 3D human pose from monocular image. This method relies on an overcomplete auto-encoder to learn a high-dimensional latent pose representation and using CNN to map pose into learned latent space. Drawback of the Tekin's machine learning-based method is requirement of a huge manual annotation dataset and computational complexity.

Chen et al. [5] seeks to estimate and track human upper-body pose from an image or a video sequence, they used prior data-driven models typically learned from 3D motion capture data which are expensive and time-consuming to collect. Besides, only upper body pose result is their main disadvantages compare to other full body pose reconstruction models. Recently, Akhter et al. [4] published a Pose-angle Limits model for 3D human pose reconstruction which is state-of-the-art outperforms existing. This method trains a Pose-angle Limits model that adaptively constraints joints angle to eliminate abnormal poses based on prior knowledge. However, this model still processes frame-by-frame independently while all frames in a sequence are highly correlated. In this paper, we consider [4] as the baseline, more details on Section 4.

This paper emphasizes the novel idea of taking advantage of inter-frames temporal relationship in a sequence of poses. Given a 2D human pose sequence, we firstly use Orthogonal Matching Pursuit algorithm and pre-trained Pose-angle Limits model to estimate prior 3D pose results. This pre-trained pose-dependent model were introduced by [3] that solve the fixed joint angle problems. Secondly, temporal models independently correct and smooth sequence of 3D poses to generate final 3D poses. Our method have been tested on CMU dataset and recorded Vietnamese Traditional Dance sequences to show state-of-the-art reconstruction results compare to other existing methods.

## 3. Proposed Methods

### 3.1 Framework Overview

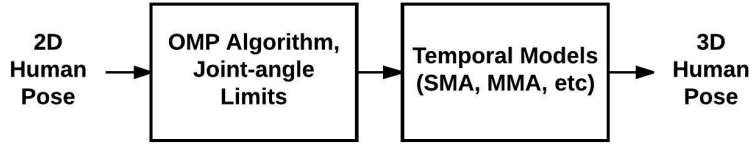

Figure 1: Overview of our framework

The outline of our 3D human pose reconstruction from 2D human pose framework is shown in Figure 1. Given a 2D skeleton (15 joints model), we firstly use a greedy Orthogonal Matching Pursuit (OMP) algorithm [3] to estimate 3D poses that minimizes projection errors. However, there always exist a set of satisfied 3D poses corresponding to a single 2D pose. In this research, we use a pre-trained Pose-angle Limits model to eliminate invalid poses and choose the best candidate. Invalid poses stand for abnormal and impossible joint angles which was introduced by [4]. Now, considering a sequence of 2D skeleton, the second step leverages temporal relationship between consecutive frames to reduce high frequency pose movements by applying temporal models which are explained detail in Subsection 3.3. Finally, the framework outputs a smooth sequence of 3D human poses.

### 3.2 Pose-angle Limits model and OMP Algorithm

We firstly re-cap the concept of Pose-angle Limits model which was originally introduced by [4]. In 2015, Akhter et al. captured a dataset of motions where actors were asked to keep their upper-arm fixed, fully flex and extend their lower-arms and then turn them inwards to outwards. Similar motions were recorded for the legs. From this data, a Pose-angle Limits model of kinematic skeleton was trained. Given a skeleton of P joints: $X = [X_1^T, \ldots, X_P^T]^T \in R^{3P \times 1}$, Pose-angle Limit provides a function to determine either an joint is in a valid pose or not. We re-use this pre-trained Pose-angle Limit in our framework.

$$isvalid(X): R^{3 \times P} \to \{0,1\}^P. \quad\quad 1$$

In [3], a linear combination of basis poses are used to represent a sparse form of 3D human pose:

$$X = \mu + \Sigma_{i=1}^{K} w_i b_i = \mu + B^* w, \quad\quad 2$$
$$\{b_i\}_{i \in I_{B^*}} \in B^* \subset \hat{B}.$$

where $\mu$ is mean-pose obtained by averaging poses from the CMU motion capture dataset; $K$ is the number of basis poses; $b_i$ are the basis poses, $w$ is a vector of pose coefficients $w_i$; matrix $B^*$ is a column-wise concatenation of basis poses $b_i$ selected with column indices $I_{B^*}$; $\hat{B}$ is an over-complete dictionary of basis components; $B^*$ is optimal subset from $\hat{B}$, contains a number of basis pose represent $X$.

In order to find the unknown 3D pose X, they minimize projection error to find $w, I_{B^*}, s$ (camera scale) and R (camera rotation) using Orthogonal Matching Pursuit (OMP) algorithm subject to anthropometric regularization:

$$\quad\quad 3$$
$$C_r(w, I_{B^*}, s, R) = \left\| x - s(I \otimes R_{1:2})(B^* w + \mu) \right\|_2^2.$$

where $\otimes$ denotes Kronecker product, $R_{1:2}$ gives first two rows of matrix R. Once $w, I_{B^*}, s$ are known, the 3D pose is estimated as:

$$X = (I \otimes R)(B^* w + \mu). \quad\quad 4$$

Pose-angle Limits model and OMP algorithm are integrated in the first stage of our framework to estimate a single 3D human pose for each 2D pose input frame. While OMP algorithm tends to minimize projection error, Pose-angle Limits model try to eliminate invalid poses based on pre-trained 3D pose dataset.

## 3.3 Temporal Models

Many approaches have been proposed on topics of 3D human pose reconstruction, but they almost focus on single frame, in other words, only intra-frame relationship is used. In fact, borrowing idea from video coding, it is obvious that consecutive frames always have a high correlation, then consecutive 3D skeletons should also have a high correlation. From this point of view, we propose a novel approach using temporal models to utilize the inter-frame relationship. Those temporal models are not only reduce the high frequency joint movements, but also smooth the actions.

Given a sequence of N frames 3D skeleton (15 joints model), we consider each joint 3D coordinate $(x, y, z)$ as discrete-time signals S.

$$S = (S_1, S_2, \ldots, S_N). \qquad 5$$

In order to smooth sequence of 3D skeleton, we basically apply temporal models to all the discrete-time signals S length N. In this study, four temporal models are implemented, including Simple, Exponential, Weighted and Modified Moving Average.

### 3.3.1 Simple Moving Average

The most basic Simple Moving Average (SMA) is averaging signal from the left most period with a chosen sliding window size $w$. SMA is mathematically simple, but it efficiently removes high frequency data. The bigger number of $w$, the smoother the SMA.

$$S_t^{sma} = \begin{cases} S_t, & t \in [1, w] \\ \dfrac{S_t + S_{t-1} + \cdots + S_{t-w+1}}{w}, & t \in [w+1, N]. \end{cases} \qquad 6$$

where
S indicates a length N signal;
$w$ is a pre-defined window size.

### 3.3.2 Exponential Moving Average

Exponential moving averages are another form of weighted averaging. EMA allows user control data smoothness by adjusting either window size parameter w or smoothing factor $\alpha \in [0,1]$. In this paper, we use smooth factor of $\alpha = \dfrac{2}{w+1}$, following formula below:

$$S_t^{ema} = \begin{cases} S_t, & t = 1 \\ (1-\alpha)S_{t-1}^{ema} + \alpha S_t, & t \in (1, N]. \end{cases} \qquad 7$$

where
S indicates a length N signal;
$w$ is an user-defined window size;
$\alpha = \dfrac{2}{w+1}$ is smoothing factor.

### 3.3.3 Weighted Moving Average

Weighted moving average mathematically is the convolution between data series and a fixed weighting function. Basically, weighting function give current data more weight than older data to reduce proportionally the implication of previous data points. In this paper, we use weighted function as below:

$$S_t^{wma} = \begin{cases} S_t, & t \in [1, w] \\ \dfrac{wS_t + (w-1)S_{t-1} + \cdots + 2S_{t-(w-2)} + 1S_{t-(w-1)}}{w + (w-1) + (w-2) + \cdots + 2 + 1}, & t \in [w+1, N]. \end{cases} \quad 8$$

where
S indicates a length N signal;
$w$ is an user-defined window size.

### 3.3.4 Modified Moving Average

Modified moving average is a short form of exponential moving average, with $\alpha = \frac{1}{w}$.

$$S_t^{mma} = \begin{cases} S_t, & t = 1 \\ (1 - \dfrac{1}{w})S_{t-1}^{mma} + \dfrac{1}{w}S_t, & t \in (1, N]. \end{cases} \quad 9$$

where
S indicates a length N signal;
$w$ is an use-defined window size.

## 4 Experiments

We demonstrate the performance of our framework in 3D human pose reconstruction in three experiments. First, we compare our method with state-of-the art methods running on CMU dataset. Secondly, we test the normal distribution (Gaussian) noise sensitivity by observing average error versus signal-to-noise ratio. Another test uses a Vietnamese Traditional Dace dataset to indicate the smoothness of our reconstructed pose sequence. Basically, our proposed method outperform existing works with lower Euclidean reconstruction error and producing naturally smooth 3D human pose sequence.

### 4.1 Quantitative Evaluation on CMU Dataset

### CMU data acquisition

We use a subset of 11 sequences (approximate 25 thousands of frames) from the CMU Motion Capture Database which was introduced by Carnegie Mellon University [12]. The subset is then normalized in several steps. First, we retarget subset motions into a unique skeleton because of CMU was performed by multiple actors. As a consequence, all retargeted motion have the same bone lengths. Secondly, we convert original rotational skeleton into 3D positional data, selecting 17 major joints, then translating local origin of the skeleton (hip joint) to global origin.

### Euclidean comparison metric

We use obtained CMU 3D poses as ground-truth, named $S^{gtr}$. Beside, a camera model has been implemented to orthogonally project 3D skeleton onto a 2D plane, produces images of corresponding 2D poses, named $S^{2d}$. Then, we feed projected 2D poses to 3D pose reconstruction frameworks, output estimated 3D poses, called $S^{rec}$. For comparison, we calculate average Euclidean distance between reconstructed poses and ground-truth poses correspondingly, as shown in Table 1 and Figure 2.

Additionally, ground-truth $S^{gtr}$ and Reconstruction $S^{rec}$ are both tend to represent a 3D pose, but they are likely to have different coordinate because of different camera parameter models. In order to overcome this problem, we implemented a Proscrustes Alignment (PA) which determines a linear transformation (combination of translation, reflection, orthogonal rotation, and scaling) [13] to conform $S^{rec}$ to $S^{gtr}$ as small dissimilarity as possible.

$$(PA_{S^{rec}}, PA_{S^{gtr}}) = PA_{(S^{rec}, S^{gtr})},$$
$$Err = Mean(D_{Eclidean}(PA_{S^{rec}}, PA_{S^{gtr}})).$$

## Results on CMU dataset

In this paper, we consider the work of Akhter et al. [4] as the baseline. Table 1 numerically depicts joint-by-joint reconstruction error percentages of our temporal frameworks compared to baseline method [4]. The lower percentage of error, the better the method is. In total, there are four trials including varieties of temporal models: SMA, EMA, WMA and MMA. It is clearly that temporal models are useful to smooth the pose sequences and reduce reconstruction errors while four temporal models perform better than the baseline. In average, our framework using Modified Moving Average (MMA) model have lowest error percentage, about ten percent lower than baseline, as shown in Table 1. Additionally, Figure 2 specifically indicates joint-by-joint Euclidean reconstruction errors of MMA and baseline model visually.

| Joint | Baseline | Our SMA | Our EMA | Our WMA | Our MMA |
|---|---|---|---|---|---|
| Hip | 100.00% | 84.93% | 84.50% | 85.78% | **81.33%** |
| Neck | 100.00% | 86.65% | 85.34% | 87.59% | **82.83%** |
| L-Shoulder | 100.00% | 85.57% | 84.63% | 86.29% | **81.78%** |
| L-Elbow | 100.00% | 84.18% | 83.53% | 85.01% | **80.54%** |
| L-Hand | 100.00% | 94.07% | 93.65% | 94.35% | **92.74%** |
| R-Shoulder | 100.00% | 89.12% | **85.67%** | 89.80% | 86.82% |
| R-Elbow | 100.00% | 86.60% | 85.83% | 87.04% | **82.93%** |
| R-Hand | 100.00% | 88.79% | 88.25% | 89.22% | **85.53%** |
| Head | 100.00% | 88.35% | 87.25% | 88.98% | **84.50%** |
| R-Hip | 100.00% | 96.86% | 96.66% | 97.03% | **95.43%** |
| R-Knee | 100.00% | 100.56% | 100.22% | 100.79% | **99.87%** |
| R-Foot | 100.00% | 102.11% | **98.27%** | 102.18% | 99.13% |
| L-Shoulder | 100.00% | 94.80% | 94.50% | 94.99% | **93.63%** |
| L-Knee | 100.00% | 100.06% | 99.99% | 100.19% | **99.08%** |
| L-Foot | 100.00% | 100.06% | 100.11% | **99.81%** | 98.87% |
| **AVERAGE** | **100.00%** | **92.18%** | **91.23%** | **92.60%** | **89.67%** |

Table 1: Reconstruction error percentages comparison joint-by-joint.

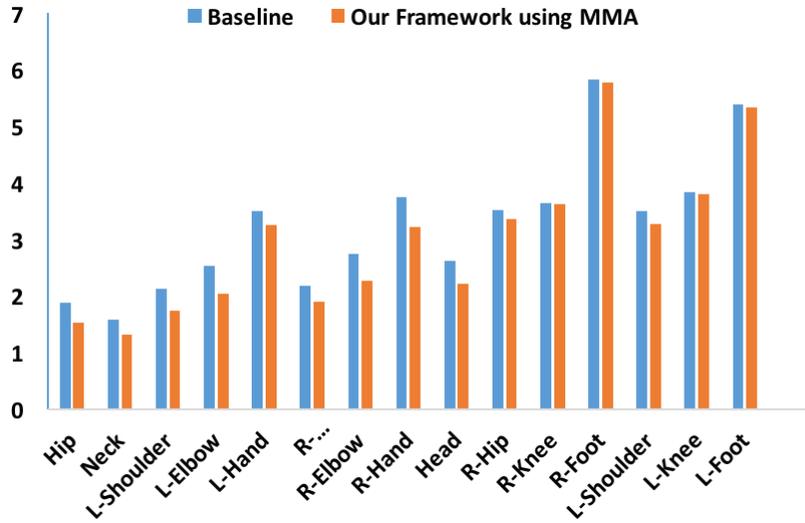

Figure 2: Euclidean reconstruction error of our framework using MMA model vs. baseline.

It is important to mention the smoothness of estimated 3D pose sequences. Due to our temporal model takes advantage of inter-frame relationship, reconstructed pose sequence clearly look more natural and smooth than others methods, as shown in Figure 3 and Figure 6.

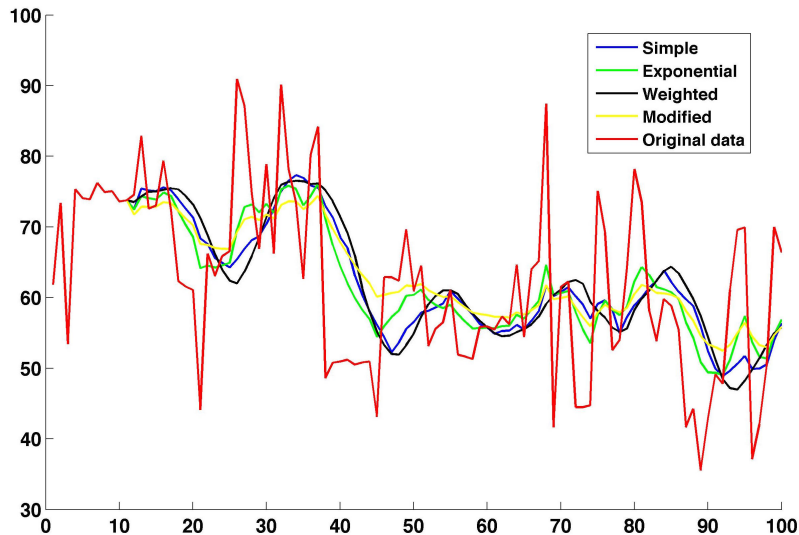

Figure 3: Signal smoothing visualization.

## 4.2 Gaussian Noise Sensitivity

In this experiment, we add Gaussian noise on 2D poses input to check method's sensitivity against noise. Then we observe the change of reconstruction error versus signal-to-noise ratio (SNR in dB). The bigger the SNR, the noisier of the signal and the lower reconstruction error, the better the method against Gaussian noise. Figure 4 visualizes an original signal and the added noises respectively by SNR={ 1, 9, 17} dB. Figure 5 shows the percentage error of our framework versus Akhter et al. [4]. It is clear that our method have lower reconstruction error, outperforms others in terms of sensitivity against Gaussian noise.

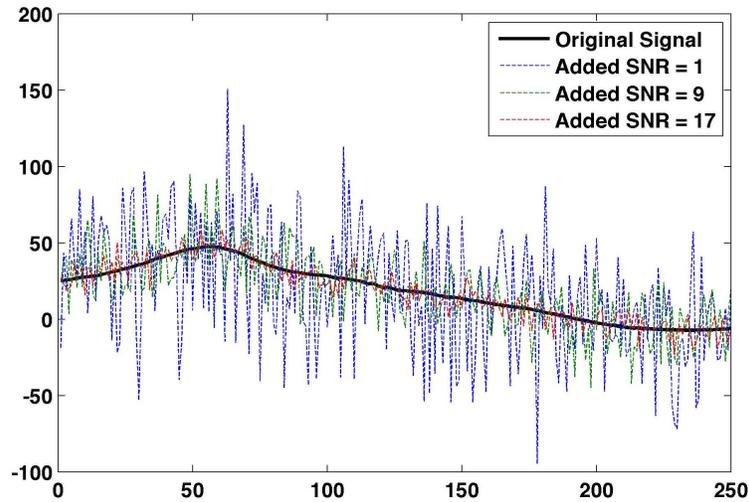

Figure 4: Visualization of noisy signals.

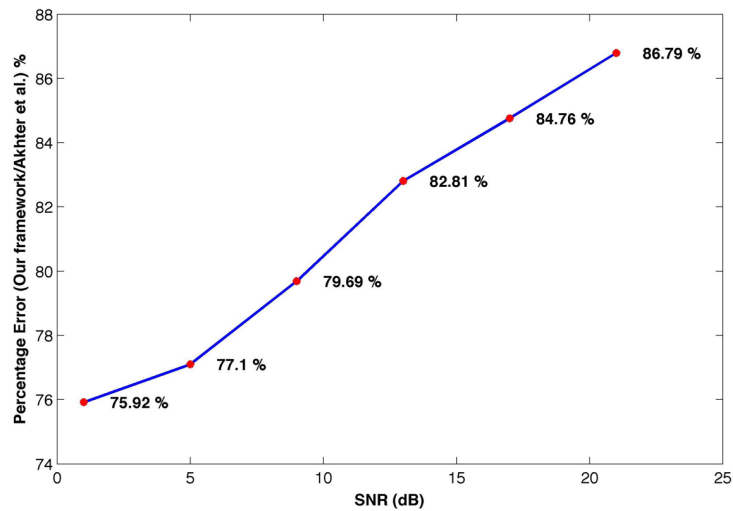

Figure 5: The change of reconstruction error versus Gaussian SNR

## 4.3  Result on Apsara Vietnamese Traditional Dance Dataset

In terms of practical applications, our framework have chances to be tested in a cultural heritage preservation project which attempts to protect Vietnamese traditional dances using computer vision techniques. This work targets to store, synthesize and render traditional dances not only in single-view video, but also in corresponding 3D human poses.

In details, our group manually record Apsara's performances (a Vietnamese traditional dance) in format of 2D videos. Then, we use YOLO [14] and Deepcut [9] to estimate 2D actor's poses from 2D recorded videos. YOLO and Deepcut are two state-of-the-art machine learning-based methods for character segmentation and 2D pose estimation. Next step, our framework is applied to reconstruct 3D poses from 2D poses sequences. Figure 6 the process from 2D video to 3D poses. Our dance's reconstructed 3D pose sequences are reasonable and smooth enough for later synthesis and rendering.

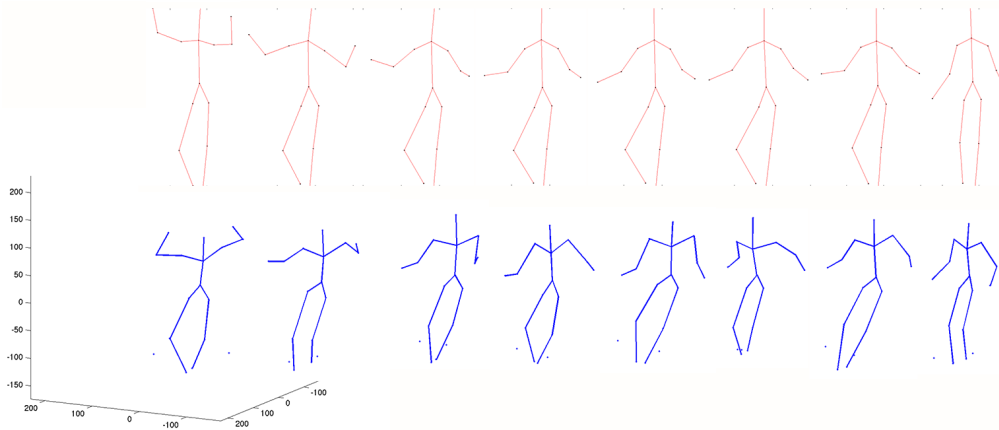

Figure 6: Apsara dance 3D pose reconstruction: Upper-parts are 2D poses; Lower-parts are corresponding 3D poses.

## 5 Conclusion

In this paper, we propose a spatial-temporal 3D human pose reconstruction framework. The key idea of our approach is combining intra and inter frames relationships in a consecutive pose sequences. Orthogonal Matching Pursuit (OMP) algorithm, pre-trained Pose-angle Limits and temporal models have been used to estimate 3D poses from 2D poses. Additionally, temporal models can be re-used independently as an post-processing step to smooth reconstructed motion sequences in other methods. Experiment results on CMU dataset and Vietnamese Traditional Dance clearly show that our framework outperform existing methods in terms of Euclidean reconstruction error. In details, using Modified Moving Average model, our framework have lower reconstruction error (about 10 percent) compare to the baseline [4]. On the other hand, our method have better sensitivity of dealing with Gaussian noise. It is also important to mention that our 3D pose sequences are more smooth and realistic than others.

## Acknowledgement

This research is supported/funded by EU project "Multimedia Application Tools for Intangible Cultural Heritage Conservation and Promotion" - H2020-MSCA-RISE ANIAGE (691215) and Ministry of Science and Technology Vietnam (ĐTĐL.CN-34/16).

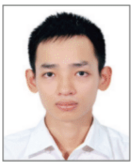

**Xuan Thanh Nguyen**/ http://orcid.org/0000-0001-5464-0327

He is Ph.D. student at ESIEE Paris, working on mathematical morphology, computer vision and machine learning. He received B.S. in Computer Science from University of Engineering and Technology, Vietnam National University (UET, VNU), Hanoi, Vietnam in 2013. He got M.Sc at JAIST-Japan in 2016.

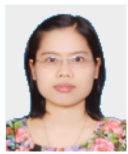

**Thi Duyen Ngo/** https://orcid.org/0000-0002-1557-9153

Thi Duyen Ngo received her Bachelor degree in Information Technology in 2005 from University of Engineering and Technology (UET), Vietnam National University – Hanoi (VNUH), where she has been working as a lecturer since 2006. She received a Ph.D. at UET, VNUH in 2016. Her research interests are speech processing and computer vision.

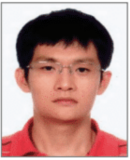

**Thanh Ha Le**/ http://orcid.org/0000-0002-7288-0444

Ha Le Thanh received B.S. and M.S. degrees in Information Technology from the College of Technology, Vietnam National University, Hanoi in 2005. He received a Ph.D. at the Department of Electronics Engineering at Korea University. In 2010, he joined the Faculty of Information Technology, University of Engineering and Technology, Vietnam National University, Hanoi as an associate professor. His research interests are multimedia processing, coding satellite image processing and computer vision.